# A New Technique for Combining Multiple Classifiers using The Dempster-Shafer Theory of Evidence

**Ahmed Al-Ani**                                    A.ALANI@QUT.EDU.AU
**Mohamed Deriche**                                 M.DERICHE@QUT.EDU.AU
*Signal Processing Research Centre*
*Queensland University of Technology*
*GPO Box 2434, Brisbane, Q 4001, Australia*

## Abstract

This paper presents a new classifier combination technique based on the Dempster-Shafer theory of evidence. The Dempster-Shafer theory of evidence is a powerful method for combining measures of evidence from different classifiers. However, since each of the available methods that estimates the evidence of classifiers has its own limitations, we propose here a new implementation which adapts to training data so that the overall mean square error is minimized. The proposed technique is shown to outperform most available classifier combination methods when tested on three different classification problems.

## 1. Introduction

In the field of pattern recognition, the main objective is to achieve the highest possible classification accuracy. To attain this objective, researchers, throughout the past few decades, have developed numerous systems working with different features depending upon the application of interest. These features are extracted from data and can be of different types like continuous variables, binary values, etc. As such, a classification algorithm used with a specific set of features may not be appropriate with a different set of features. In addition, classification algorithms are different in their theories, and hence achieve different degrees of success for different applications. Even though, a specific feature set used with a specific classifier might achieve better results than those obtained using another feature set and/or classification scheme, we can not conclude that this set and this classification scheme achieve the best possible classification results (Kittler, Hatef, Duin, & Matas, 1998). As different classifiers may offer complementary information about the patterns to be classified, combining classifiers, *in an efficient way*, can achieve better classification results than any single classifier (even the best one).

As explained by Xu et al. (1992), the problem of combining multiple classifiers consists of two parts. The first part, closely dependent on specific applications, includes the problems of "How many and what type of classifiers should be used for a specific application?, and for each classifier what type of features should we use?", as well as other problems that are related to the construction of those individual and complementary classifiers. The second part, which is common to various applications, includes the problems related to the question "How to combine the results from different existing classifiers so that a better result can be obtained?". In our work, we will be concentrating on problems related to the second issue.

The output information from various classification algorithms can be categorized into three levels: the abstract, the rank, and the measurement levels. In the abstract level, a classifier only outputs a unique label, as in the case of syntactic classifiers. For the rank level, a classifier ranks all labels or a subset of the labels in a queue with the label at the top being the first choice. This type was discussed by Ho et al. (1994). For the measurement level, a classifier attributes to each class a measurement value that reflects the degree of confidence that a specific input belongs to a given class. Among the three levels, the measurement level contains the highest amount of information while the abstract level contains the lowest. For this reason, we adopted, in this work, the measurement level.

Kittler et al. (1998) differentiated between two classifier combination scenarios. In the first scenario, all the classifiers use the same representation of the input pattern. On the other hand, each classifier uses its own representation of the input pattern in the second scenario. They illustrated that in the first case, each classifier can be considered to produce an estimate of the same a posteriori class probability. However, in the second case it is no longer possible to consider the computed a posteriori probabilities to be estimates of the same functional value, as the classification systems operate in different measurement systems. Kittler et al. (1998) focused on the second scenario, and they conducted a comparative study of the performance of several combination schemes namely; product, sum, min, max, and median. By assuming the joint probability distributions to be conditionally independent, they found that the sum rule gave the best results. A well known approach that has been used in combining the results of different classifiers is the weighted sum, where the weights are determined through a Bayesian decision rule (Lam & Suen, 1995). An alternative method was presented by Hashem & Schmeiser (1995), where a cost function was used to minimize the mean square error (MSE) in order to calculate a linear combination of the corresponding outputs from a number of trained artificial neural networks (ANNs). The expectation maximization algorithm was used by Chen & Chi (1998) to perform the linear combination. The fuzzy integral has been used by Cho & Kim (1995a, 1995b) to combine multiple ANNs, while (Rogova, 1994; Mandler & Schurmann, 1988) have used the Dempster-Shafer theory of evidence to combine the result of several ANNs. Many other combination methods have also been used to combine classifiers, such as bagging and boosting (Dietterich, 1999), which are powerful methods for diversifying and combining classification results obtained using a single classification algorithm and a specific feature set. In bagging, we get a family of classifiers by training on different portions of the training set. The method works as follows. We first create $N$ training bags. A single training bag is obtained by taking a training set of size $S$ and sampling this training set $S$ times with replacement. Some training instances will occur multiple times in a bag, while others may not appear at all. Next, each bag is used to train a classifier. These classifiers are then combined. Boosting, on the other hand, is based on multiple learning iterations. At each iteration, instances that are incorrectly classified are given a greater weight in the next iteration. By doing so, in each iteration, the classifier is forced to concentrate on instances it was unable to correctly classify in earlier iterations. In the end, all of the trained classifiers are combined.

In this paper, we will focus on combining classification results obtained using $N$ different feature sets, $\mathbf{f}^1, \cdots, \mathbf{f}^N$. Each feature set will be used to train a classifier, and hence there will be $N$ different classifiers, $c^1, \cdots, c^N$. For a specific input $\mathbf{x}$, each classifier $c^n$ produces a real vector $\mathbf{y}^n = [y^n(1), \cdots y^n(k), \cdots y^n(K)]^T$, where $K$ is the number of class labels and $y^n(k)$ corresponds to the degree that $c^n$ considers $\mathbf{x}$ has the label $k$. This degree could be a probability, as in the Bayesian classifier, or any other scoring system. Fig. 1 shows the block diagram of a multi-classifier recognition system.

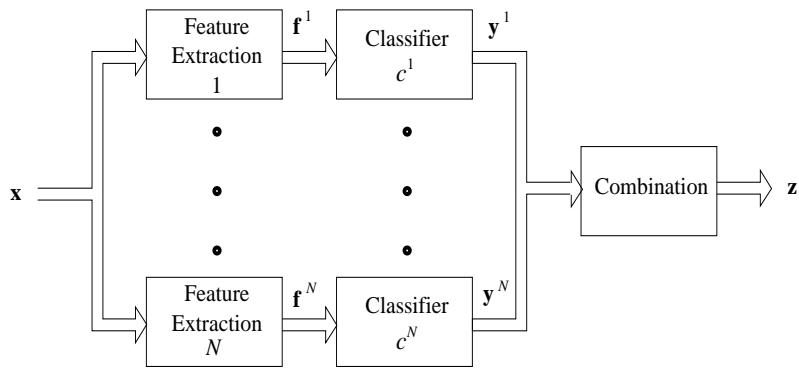

Figure 1: A multi-classifier recognition system

Unlike statistical-based combination techniques, the Dempster-Shafer theory of evidence has the ability to represent uncertainties and lack of knowledge. This is quite important for the problem of classifier combination, because there is usually a certain level of uncertainty associated with the performance of each of the classifiers. Since available classifier combination methods based on this theory do not accurately estimate the evidence of classifiers, this paper attempts to solve this issue by proposing a new technique based on the gradient descent learning algorithm, which aims at minimizing the MSE between the combined output and the target output of a given training set. Aha (1995) gave the following definition for learning:

> Learning denotes changes in the system that are adaptive in the sense that they enable the system to do the same task or tasks drawn from the same population more effectively the next time.

Based on the above, we show that instead of attempting to find an analytical formula which accurately measures evidence, one can obtain a very good estimate of evidence by just using appropriate learning procedures, as will be discussed later.

Some basic concepts of the Dempster-Shafer theory of evidence are presented in the next section. Section three discusses the existing methods for computing evidence. The proposed combination technique is presented in section four. Section five compares the proposed algorithm to other conventional methods used by Kittler et al. (1998), the fuzzy integral, and a previous implementation of the Dempster-Shafer theory. Section six provides a conclusion to the paper.

## 2. The Dempster-Shafer Theory of Evidence

The Dempster-Shafer (D-S) theory of evidence (Shafer, 1976) is a powerful tool for representing uncertain knowledge. This theory has inspired many researchers to investigate different aspects related to uncertainty and lack of knowledge and their applications to real life problem. Today, the D-S theory covers several different models, such as the theory of hints (Kohlas & Monney, 1995) and the transferable belief model (TBM) (Smets, 1998). The latter will be adopted in this paper as it represents a powerful tool for combining measures of evidence.

Let $\Theta = \{\theta_1, \ldots, \theta_K\}$ be a finite set of possible hypotheses. This set is referred to as the *frame of discernment*, and its powerset denoted by $2^\Theta$. Following are the basic concepts of the theory:

**Basic belief assignment (BBA)**. A basic belief assignment $m$ is a function that assigns a value in $[0, 1]$ to every subset $\mathcal{A}$ of $\Theta$ and satisfies the following:

$$m(\emptyset) = 0, \text{ and } \sum_{\mathcal{A} \subseteq \Theta} m(\mathcal{A}) = 1 \tag{1}$$

It is worth mentioning that $m(\emptyset)$ could be positive when considering unnormalized combination rule as will be explained later. While in probability theory a measure of *probability* is assigned to atomic hypotheses $\theta_i$, $m(\mathcal{A})$ is the part of *belief* that supports $\mathcal{A}$, but does not support anything more specific, *i.e.*, strict subsets of $\mathcal{A}$. For $\mathcal{A} \neq \theta_i$, $m(\mathcal{A})$ reflects some ignorance because it is a belief that we cannot subdivide into finer subsets. $m(\mathcal{A})$ is a measure of support we are willing to assign to a composite hypothesis $\mathcal{A}$ at the expense of support $m(\theta_i)$ of atomic hypotheses $\theta_i$. A subset $\mathcal{A}$ for which $m(\mathcal{A}) > 0$ is called a *focal element*. The *partial ignorant* associated with $\mathcal{A}$ leads to the following inequality: $m(\mathcal{A}) + m(\overline{\mathcal{A}}) \leq 1$, where $\overline{\mathcal{A}}$ is the compliment of $\mathcal{A}$. In other words, the D-S theory of evidence allows us to represent *only our actual knowledge* without being forced to overcommit when we are ignorant.

**Belief function.** The belief function, $bel(.)$, associated with the BBA $m(.)$ is a function that assigns a value in $[0, 1]$ to every nonempty subset $\mathcal{B}$ of $\Theta$. It is called "*degree of belief in $\mathcal{B}$*" and is defined by

$$bel(\mathcal{B}) = \sum_{\mathcal{A} \subseteq \mathcal{B}} m(\mathcal{A}) \tag{2}$$

We can consider a basic belief assignment as a generalization of a probability density function whereas a belief function is a generalization of a probability function.

**Combination rule.** Consider two BBAs $m_1(.)$ and $m_2(.)$ for belief functions $bel_1(.)$ and $bel_2(.)$ respectively. Let $\mathcal{A}_j$ and $\mathcal{B}_k$ be focal elements of $bel_1$ and $bel_2$ respectively. Then $m_1(.)$ and $m_2(.)$ can be combined to obtain the belief mass committed to $\mathcal{C} \subset \Theta$ according to the following combination or *orthogonal sum* formula (Shafer, 1976),

$$m(\mathcal{C}) = m_1 \oplus m_2(\mathcal{C}) = \frac{\displaystyle\sum_{j,k, \mathcal{A}_j \cap \mathcal{B}_k = \mathcal{C}} m_1(\mathcal{A}_j) m_2(\mathcal{B}_k)}{1 - \displaystyle\sum_{j,k, \mathcal{A}_j \cap \mathcal{B}_k = \emptyset} m_1(\mathcal{A}_j) m_2(\mathcal{B}_k)}, \quad \mathcal{C} \neq \emptyset \tag{3}$$

The denominator is a normalizing factor, which intuitively measures how much $m_1(.)$ and $m_2(.)$ are conflicting. Smets (1990) proposed the unnormalized combination rule:

$$m_1 \textcircled{\cap} m_2(\mathcal{C}) = \sum_{\mathcal{A}_j \cap \mathcal{B}_k = \mathcal{C}} m_1(\mathcal{A}_j) m_2(\mathcal{B}_k), \quad \forall \mathcal{C} \subseteq \Theta \qquad (4)$$

This rule implies that $m(\emptyset)$ could be positive, and in such case reflects some kind of contradiction in the belief state. In this work we will consider that $m(\emptyset) = 0$ and use the normalized combination rule. A comparison between normalized and unnormalized combination rules for the problem of combining classifiers will be considered in the future.

**Combining several belief functions.** The combination rule can be easily extended to several belief functions by repeating the rule for new belief functions. Thus the pairwise orthogonal sum of $n$ belief functions $bel_1, bel_2, \cdots, bel_n$, can be formed as

$$((bel_1 \oplus bel_2) \oplus bel_3) \cdots \oplus bel_n = \bigoplus_{i=1}^{n} bel_i \qquad (5)$$

**Notation.** According to Smets (2000), the full notation for $bel$ and its related functions is:

$$bel_{Y,t}^{\Theta \Re}[EC_{Y,t}](w_0 \in \mathcal{A}) = x$$

where $Y$ represents the agent, $t$ the time, $\Theta$ the frame of discernment, $\Re$ a boolean algebra of subsets of $\Theta$, $w_0$ the actual world, $\mathcal{A}$ a subset of $\Theta$, and $EC_{Y,t}$ all what agent $Y$ knows at $t$. Thus, the above expression denotes that the degree of belief held by $Y$ at $t$ that $w_0$ belongs to the set $\mathcal{A}$ of worlds is equal to $x$. The belief is based on the evidential corpus $EC_{Y,t}$ held be $Y$ at $t$.

In practice, many indices can be omitted for simplicity sake. Usually $\Re$ is the power set of $\Theta$, which is $2^\Theta$. When $bel$ is defined on $2^\Theta$, $\Re$ is not explicitly stated. '$w_0 \in \mathcal{A}$' is denoted as '$\mathcal{A}$'. $Y$ and/or $t$ are omitted when the values of the missing elements are clearly defined from the context. Furthermore, $EC$ is usually just a conditioning event. So, $bel(\mathcal{A})$ is one of the most often used notations (Smets, 2000). In the proposed method, we will adopt the following notation: $bel_n(\theta_k)$, where the agent is the classifier, and the subsets of concern are the class labels.

It is important to mention that the combination rule given by Eq. 3 assumes that the belief functions to be combined are independent. Consider that we have certain information and would like to measure its belief, then we can think of this process as a mapping from the "original information level" to the "belief level". Liu & Bundy (1992) explained that independence in the original information level would lead to independence in the belief level. But, if two independent belief functions are rooted to the original information level, then their original information may or may not be independent. For the problem of combining multiple classifiers, the original information level consists of outputs of the classifiers to be combined, while the belief level consists of the evidence of these classifiers (or their BBAs). The assumption that these BBAs are independent, whether obtained from independent or dependent original information, can hence justify the use of D-S theory. In fact, many existing classifier combination methods assume the classification results of different classifiers to be independent (Mandler & Schurmann, 1988; Hansen & Salamon, 1990; Xu et al., 1992). Since the classifiers' evidence plays a crucial role in the combination performance, there is an increased interest in the proper estimation of such evidence. In the next section, we discuss how a number of existing classifier combination methods estimate evidence of classifiers, and in section 4 we present our proposed method.

## 3. Existing Methods for Computing Evidence

Mandler & Schurmann (1988) proposed a method that transforms distance measures of the different classifiers into evidence. This was achieved by first calculating a distance between learning data sets and a number of reference points in order to estimate statistical distributions of intra- and interclass distances. For both, the a posteriori probability function was estimated, indicating to which degree an input pattern belongs to a certain reference point. Then, for each class label, the class conditional probabilities were combined into evidence value ranging between 0 and 1, which was considered as the BBA of that class. Finally, Dempster's combination rule was used to combine the BBAs of the different classifiers to give the final result. As explained by Rogova (1994), this method brought forward questions about the choice of reference vectors and the distance measure. Moreover, approximations associated with estimation of parameters of statistical models for intra- and interclass distances can lead to inaccurate measure of the evidence.

Xu et al. (1992) used $K + 1$ classes to perform the classification task, where for the $(K + 1)^{th}$ class denotes that the classifier has no idea about which class the input comes from. For each classifier $c^n$, $n = 1..N$, recognition, substitution, and rejection rates ($\epsilon_r^n$, $\epsilon_s^n$, and $1 - \epsilon_r^n - \epsilon_s^n$) were used as a measure of BBA, $m_n$, on $\Theta$ as follows:

1. If the maximum output of a specific classifier belongs to $K + 1$, then $m_n$ has only a focal element $\Theta$ with $m_n(\Theta) = 1$.

2. When the maximum output belongs to one of the $K$ classes, $m_n$ has two focal elements $\theta_k$ and $\overline{\theta}_k$ with $m_n(\theta_k) = \epsilon_r^n$, $m_n(\overline{\theta}_k) = \epsilon_s^n$. As the classifier says nothing about any other propositions, $m_n(\Theta) = 1 - m_n(\theta_k) - m_n(\overline{\theta}_k)$.

The drawback of this method is again the way evidence is measured. There are two problems associated with this method. Firstly, many classifiers do not produce binary outputs, but rather probability like outputs. So, in the first case, it is inaccurate to assign 0 to both $m_n(\theta_k)$ and $m_n(\overline{\theta}_k)$. Secondly, this way of measuring evidence ignores the fact that classifiers normally do not have the same performance with different classes. This had a clear impact on the performance of this combination method when compared with other conventional methods especially the Bayesian (Xu et al., 1992).

Rogova (1994) used several proximity measures between a reference vector and a classifier's output vector. The proximity measure that gives the highest classification accuracy was later transformed into evidences. The reference vector used was the mean vector, $\mu_k^n$, of the output set of each classifier $c^n$ and each class label $k$. A number of proximity measures, $d_k^n$, for $\mu_k^n$ and $\mathbf{y}^n$ were considered. For each classifier, the proximity measure of each class is transformed into the following BBAs:

$$
\begin{aligned}
m_k(\theta_k) &= d_k^n, \quad m_k(\Theta) = 1 - d_k^n \\
m_{\overline{k}}(\overline{\theta}_k) &= 1 - \prod_{l \neq k}(1 - d_l^n), \quad m_{\overline{k}}(\Theta) = \prod_{l \neq k}(1 - d_l^n)
\end{aligned}
$$

The evidence of classifier $c^n$ and class $k$ is obtained by combining the knowledge about $\theta_k$, thus $m_k \oplus m_{\overline{k}}$. Finally, Dempster's combination rule was used to combine evidences for all classifiers to obtain a measure of confidence for each class label. Note that the first combination was performed with respect to the class label (Rogova used the notations $k$ and $\overline{k}$), while in the second one the agent was $n$. This idea was a promising one. However, the major drawback is the way the reference vectors are calculated, where the mean of output vectors may not be the best choice. Also, trying several proximity measures and choosing the one that gives the highest classification accuracy is itself questionable.

## 4. The Proposed Combination Technique

In this section we will estimate the value of $m_n(\theta_k)$, which represents the belief in class label $k$ that is produced by classifier $c^n$. In addition, we will also estimate $m_n(\Theta)$, which reflects the ignorance associated with classifier $c^n$. Since the ultimate objective is to minimize the MSE between the combined classification results and the target output, $m_n(\theta_k)$ and $m_n(\Theta)$ will be estimated using an iterative procedure that aims at attaining this objective. We will first compare $\mathbf{y}^n$, which is the output classification vector produced by classifier $c^n$, to a reference vector, $\mathbf{w}_k^n$, and the obtained distance will be used to estimate the BBAs. These BBAs will then be combined to obtain a new output vector, $\mathbf{z}$, that represents the combined confidence in each class label. $\mathbf{w}_k^n$ will be measured such that the MSE between $\mathbf{z}$ and the target vector, $\mathbf{t}$, of a training dataset is minimized. Note that there are two indices for $\mathbf{w}_k^n$. Thus, for class label $k$, we don't only consider the value assigned to it by classifier $c^n$, but rather the whole output vector (values assigned to each class label).

Let the frame of discernment $\Theta = \{\theta_1, \cdots \theta_k, \cdots, \theta_K\}$, where $\theta_k$ is the hypothesis that the input $\mathbf{x}$ is of class $k$. Considering a BBA, $m_n$, such that $m_n(\theta_k) \geq 0$, $m_n(\Theta) = 1 - \sum_{k=1}^{K} m_n(\theta_k)$, and $m_n$ is 0 elsewhere. Let $d_n(\theta_k)$ be a distance measure and $g_n$ the unnormalized ignorance of classifier $c^n$, then $m_n(\theta_k)$ and $m_n(\Theta)$ will be estimated according to the following formulas:

$$d_n(\theta_k) = \exp(-\|\mathbf{w}_k^n - \mathbf{y}^n\|^2) \tag{6}$$

$$m_n(\theta_k) = \frac{d_n(\theta_k)}{\sum_{k=1}^{K} d_n(\theta_k) + g_n} \tag{7}$$

$$m_n(\Theta) = \frac{g_n}{\sum_{k=1}^{K} d_n(\theta_k) + g_n} \tag{8}$$

where $m_n(\theta_k)$ and $m_n(\Theta)$ are the normalized values of $d_n(\theta_k)$ and $g_n$ respectively. Similar to $\mathbf{w}_k^n$, the minimized MSE will be used to estimate $g_n$.

Evidences of all classifiers are combined according to the normalized combination rule to obtain a measure of confidence of each class label. The $k^{th}$ element of the new combined vector is given by:

$$z(k) = m(\theta_k) = m_1(\theta_k) \oplus \cdots \oplus m_N(\theta_k) = \bigoplus_{n \in N} m_n(\theta_k) \tag{9}$$

For a given classifier $c^n$, let $I = \{1 \cdots N\} \setminus \{n\}$, $m_I = \bigoplus_{i \in I} m_i$, then Eq. 9 can be written as:

$$z(k) = m_I(\theta_k) \oplus m_n(\theta_k) \tag{10}$$

where according to Eq. 3, the combination of two BBAs is:

$$m_j(\theta_k) \oplus m_l(\theta_k) = \frac{m_j(\theta_k)m_l(\theta_k) + m_j(\theta_k)m_l(\Theta) + m_j(\Theta)m_l(\theta_k)}{1 - \sum_p \sum_{\substack{q \\ q \neq p}} m_j(\theta_p)m_l(\theta_q)} \tag{11}$$

$\mathbf{w}_k^n$ and $g_n$ will be initialized randomly, then their values will be adjusted according to a training dataset so that the MSE of $\mathbf{z}$ is minimized.

$$Err = \|\mathbf{z} - \mathbf{t}\|^2 \tag{12}$$

The values of $\mathbf{w}_k^n$ and $g_n$ are adjusted according to the formulas:

$$\mathbf{w}_k^n[new] = \mathbf{w}_k^n[old] - \alpha \frac{\partial Err}{\partial \mathbf{w}_k^n[old]} \tag{13}$$

$$g_n[new] = g_n[old] - \beta \frac{\partial Err}{\partial g_n[old]} \tag{14}$$

where $\alpha$ and $\beta$ are the learning rates. The terms $\partial Err / \partial \mathbf{w}_k^n$ and $\partial Err / \partial g_n$ are derived as follows:

$$\frac{\partial Err}{\partial \mathbf{w}_k^n} = \frac{\partial Err}{\partial z(k)} \frac{\partial z(k)}{\partial m_n(\theta_k)} \frac{\partial m_n(\theta_k)}{\partial \mathbf{w}_k^n} \tag{15}$$

$$\frac{\partial Err}{g_n} = \frac{\partial Err}{\partial z(k)} \frac{\partial z(k)}{\partial m_n(\theta_k)} \frac{\partial m_n(\theta_k)}{\partial g_n} \tag{16}$$

where,

$$\frac{\partial Err}{\partial z(k)} = 2[z(k) - t(k)] \tag{17}$$

$$\frac{\partial z(k)}{\partial m_n(\theta_k)} = \left\{ \left[ 1 - \sum_p \sum_{\substack{q \\ q \neq p}} m_n(\theta_p) m_I(\theta_q) \right] [m_I(\theta_k) + m_I(\Theta)] + [m_n(\theta_k) m_I(\theta_k) + \right.$$

$$\left. m_n(\theta_k) m_I(\Theta) + m_n(\Theta) m_I(\theta_k)] \left[ \sum_{\substack{p \\ p \neq k}} m_I(\theta_p) \right] \right\} \Bigg/$$

$$\left[ 1 - \sum_p \sum_{\substack{q \\ q \neq p}} m_n(\theta_p) m_I(\theta_q) \right]^2 \tag{18}$$

$$\frac{\partial m_n(\theta_k)}{\partial \mathbf{w}_k^n} = -\frac{2 \exp(-\|\mathbf{w}_k^n - \mathbf{y}^n\|^2)[\mathbf{w}_k^n - \mathbf{y}^n][\sum_{\substack{p \\ p \neq k}} d_n(\theta_p) + g_n]}{[\sum_p d_n(\theta_p) + g_n]^2} \tag{19}$$

$$\frac{\partial m_n(\theta_k)}{\partial g_n} = -\frac{d_n(\theta_k)}{[\sum_p d_n(\theta_p) + g_n]^2} \tag{20}$$

Fig. 2 shows a flow chart of these learning procedures. It has been found that adjusting the values of $g_n$ can be achieved during the first few iterations. By continuing the training to fine-tune the values of $\mathbf{w}_k^n$ until there is no further improvement on the training set, or we reach a pre-defined maximum number of epochs[1], the result could be further enhanced. Note that the weight values are adjusted by each pattern (not batch training). We fix the value of $\beta = 10^{-6}$, while $\alpha$ is first initialized to $5 \times 10^{-4}$, and is then changed according to the value of MSE, as described in the flow chart.

Although the computational cost involved in implementing our technique is higher than that of other combination methods[2], we only need to perform training once, which can be done off-line. Then, with the optimal values of $\mathbf{w}_k^n$ and $g_n$, we can perform the on-line combination, which is comparable to other combination methods.

On the other hand, as indicated in the beginning of this section, we consider a reference vector, $\mathbf{w}_k^n$, for each class. This leads to an increase in training time as the number of classes and/or classifiers increases. An alternative is to consider only using a reference value for each class, $w_k^n$. This will save more than 50% of training time for the case of several classifiers and classes. Note that the same learning formulas are applicable by replacing $\mathbf{w}_k^n$ with $w_k^n$ and $\mathbf{y}^n$ with $y_k^n$. We will refer to these two alternative approaches as DS1 and DS2, respectively. In the following section, we will compare DS1 and DS2 with other well-known combination methods.

---

1. The maximum number of epochs is set to 50 in all experiments described in this paper
2. Training time of most of the experiments conducted in section 5 required less than 3 minutes on a conventional PC

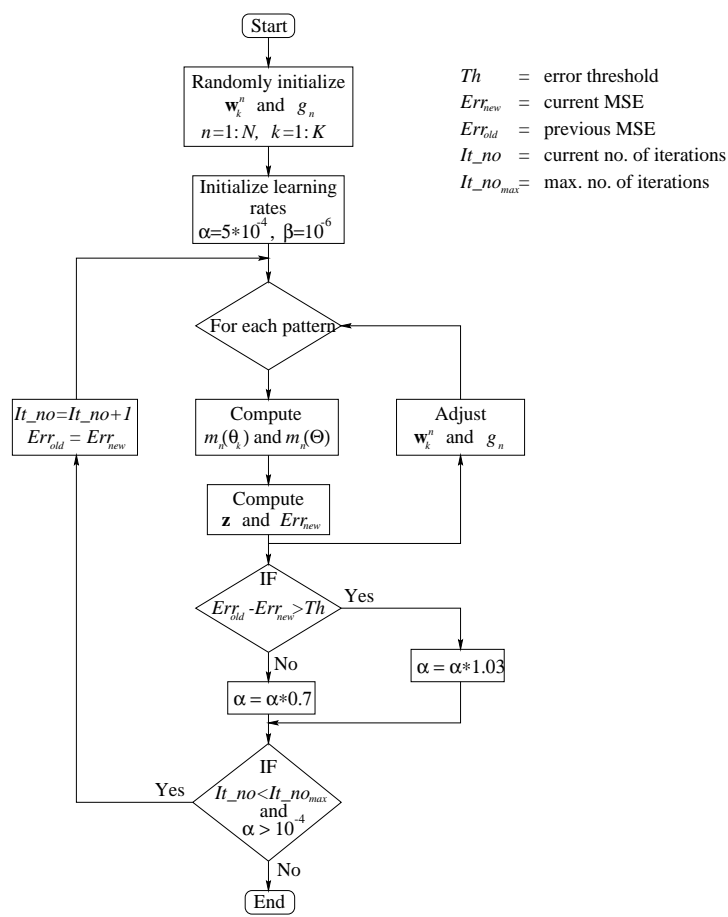

Figure 2: Training procedure of the proposed technique

It is worth mentioning that although the training procedures of both the proposed method and the backpropagation algorithm of ANN are based on minimizing the MSE using iterative approaches, the proposed method and ANN are not similar. The backpropagation training operates by passing the weighted sum of its input through an activation function, usually in a multi-layer architecture known as multi-layer perceptron (MLP). Extracting rules from a trained MLP is a very challenging problem. On the other hand, the training of the proposed method operates by measuring a distance between a classification vector and a reference vector. This distance would later be used to measure the belief of each class label for all classifiers. The final confidence of each class label is obtained by combining the beliefs of all classifiers. Unlike MLP, the belief of a given class label for each classifier indicates its contribution towards the final confidence. The reader may refer to (Denoeux, 2000) for a description of an ANN classifier based on the D-S theory.

## 5. Performance Analysis of Different Combination Methods

The following three classification problems have been considered: texture classification, classification of speech segments according to their manner of articulation, and speaker identification. ANNs are used to perform classification for the three problems. For each case, classifiers will be sorted according to their performance, such that the best classifier is referred to as $c^1$, the $2^{nd}$ best as $c^2$, and the worst one as $c^N$.

For each problem, we will consider different number of classes, and combine the results of different number of classifiers, where combining results of the best, the worst and mixtures of best and worst classifiers will be investigated. For example, if we have five classifiers and would like to combine two of these, then we will consider combining the best two, $\{c^1, c^2\}$, best one and worst one, $\{c^1, c^5\}$, and worst two classifiers, $\{c^4, c^5\}$. The following combination methods were tested: the weighted sum (WS)[3], average (Av), median (Md), maximum (Mx), majority voting (MV), fuzzy integral (FI) (Cho & Kim, 1995a)[4], Rogova's D-S method (DS0) (Rogova, 1994), and our proposed method with its two alternatives (DS1 & DS2). The training set used to train the ANNs will be used to estimate the confusion matrix for WS and FI, as well as to estimate the evidence of DS0, DS1, and DS2.

Two measures will be used to compare the performance of the different combination methods, namely: *overall performance* and *error reduction rate (ERR)*. The overall performance is the mean of classification accuracy obtained by combining all considered subsets of $2, \cdots, N$ classifiers. *ERR* is the percentage of error reduction obtained by combining classifiers with reference to the best single classifier:

$$ERR = \frac{ER_{BSC} - ER_{CC}}{ER_{BSC}} \times 100 \qquad (21)$$

where $ER_{BSC}$ is the error rate of the best single classifier and $ER_{CC}$ is the error rate obtained by combining the considered classifiers. Unlike classification accuracy, *ERR* clearly shows how the performance of the combined classifiers improves or deteriorates compared to the best single classifier. In other words, it shows the merit of performing the combination. We will specifically concentrate on the *maximum ERR* obtained by combining all the considered subsets of $2, \cdots, N$ classifiers. In addition, we will also investigate how the value of *ERR* gets affected by increasing the number of combined classifiers.

### 5.1 Texture Classification

Several experiments have been carried out for the classification of texture images. The textures considered here are: bark, brick, bubbles, leather, raffia, water, weave, wood and wool (USC, 1981). In order to obtain a better comparison between the different combination methods, we considered classifying the first two textures, then the first three, the first five and finally all the nine textures. Additive Gaussian noise, with different signal-to-noise ratio, has been added to $(1024 \times 1024)$ pixels image of each texture class to form the training and testing sets. 961 patterns were obtained from each image using $(64 \times 64)$ windows with an overlap of 32 pixels.

---

3. The weights of each classifier are determined according to the classification accuracy of each class label using the training dataset
4. The reader may refer to Appendix A for a brief description of this method

| No. of classes | $SDH_1$ | $SDH_2$ | $SDH_3$ | $SDH_4$ | $En$ |
|:---:|:---:|:---:|:---:|:---:|:---:|
| 2 | 86.96 | 85.73 | 84.44 | 85.45 | 91.14 |
| 3 | 84.58 | 84.52 | 83.91 | 86.24 | 89.72 |
| 5 | 85.10 | 84.62 | 84.34 | 83.46 | 88.84 |
| 9 | 80.97 | 77.44 | 77.51 | 75.72 | 83.65 |

Table 1: Texture classification accuracy of the five original classifiers for different number of class labels

Four nine-feature vectors were calculated using statistics of sum and difference histogram ($SDH$) of the co-occurrence matrix with different directions, vertical ($SDH_1$), horizontal ($SDH_2$), and the two diagonals ($SDH_3$ and $SDH_4$) . For each direction, the features used were: mean, variance, energy, correlation, entropy, contrast, homogeneity, cluster shade, and cluster prominence. The fractal dimension (FD) has also been used to form the tenth feature of each vector. The energy contents of texture images ($En$) has been used to form another feature vector using 9 different masks. Again the tenth feature was FD.

Each of these five feature vectors has been used as input to an ANN. The numbers of training and testing patterns depend upon number of classes considered, i.e. for the case of two classes, 15376 patterns were used to train the networks and 5766 to test them. The results obtained are shown in Table 1. Note that as the number of classes increases the overall accuracy decreases. In addition, the performance of the $En$ classifiers is found to be better than that of the other four.

| No. of classes | WS | Av | Md | Mx | MV | FI | DS0 | DS1 | DS2 |
|:---:|:---:|:---:|:---:|:---:|:---:|:---:|:---:|:---:|:---:|
| 2 | 89.16 | 89.04 | 87.66 | 90.12 | 88.09 | 90.08 | 88.70 | 90.66 | 90.72 |
| 3 | 88.52 | 88.39 | 87.41 | 88.86 | 87.30 | 88.71 | 88.40 | 90.21 | 90.08 |
| 5 | 89.60 | 89.41 | 87.99 | 89.23 | 87.83 | 90.28 | 89.52 | 92.69 | 91.50 |
| 9 | 84.96 | 84.55 | 83.37 | 82.90 | 83.23 | 86.76 | 84.87 | 89.83 | 86.79 |

Table 2: Overall performance of the various combination methods for different number of class labels (texture classification)

The overall performance of the tested combination methods for different number of class labels are shown in Table 2. For the case of 2 classes, it is clear that the overall performances of DS1 and DS2 are better than that of the other combination methods. When mixtures of good and bad classifiers are considered, the performance of combination methods, except for DS1 and DS2, is closer to or worse than that of the best single classifier. This is shown in Table 3 for the combination of $\{c^1, c^3, c^4, c^5\}$, $\{c^1, c^4, c^5\}$, $\{c^1, c^5\}$, etc[5]. When 3 and 5 classes are considered, DS1 performs slightly better than DS2, and both outperform the other methods. The gap between DS1 and other methods gets wider when all 9 classes are considered. The superiority of DS1 reflects the advantage of using the whole output vector in measuring evidences of classifiers.

---

5. The reader may refer to Appendix B for detailed results of other cases

| Classifiers | WS | Av | Md | Mx | MV | FI | DS0 | DS1 | DS2 |
|---|---|---|---|---|---|---|---|---|---|
| $c^1, c^2$ | **92.56** | **92.59** | **92.59** | **92.51** | **92.51** | **92.61** | **92.40** | 92.46 | 92.46 |
| $c^1, c^5$ | 91.16 | 91.12 | 91.12 | 91.09 | 91.09 | 91.00 | 91.33 | 91.62 | 91.61 |
| $c^4, c^5$ | 85.07 | 85.07 | 85.07 | 85.22 | 85.22 | 85.07 | 85.15 | 85.10 | 85.12 |
| $c^1, c^2, c^3$ | 91.21 | 91.21 | 88.92 | 91.62 | 88.92 | 91.69 | 90.81 | 92.40 | **92.53** |
| $c^1, c^2, c^5$ | 91.03 | 90.81 | 88.68 | 91.48 | 88.71 | 91.48 | 90.43 | **92.47** | 92.39 |
| $c^1, c^4, c^5$ | 89.80 | 89.59 | 86.21 | 91.21 | 86.21 | 91.24 | 88.88 | 91.68 | 91.78 |
| $c^3, c^4, c^5$ | 85.38 | 85.38 | 85.47 | 85.40 | 85.48 | 85.33 | 85.22 | 85.43 | 85.40 |
| $c^1, c^2, c^3, c^4$ | 89.94 | 89.70 | 87.84 | 91.47 | 89.13 | 91.59 | 89.13 | 92.21 | 92.33 |
| $c^1, c^2, c^3, c^5$ | 89.70 | 89.42 | 87.53 | 91.42 | 89.04 | 91.29 | 88.92 | 92.32 | 92.42 |
| $c^1, c^2, c^4, c^5$ | 89.72 | 89.49 | 87.37 | 91.48 | 88.94 | 91.40 | 89.00 | 92.25 | 92.26 |
| $c^1, c^3, c^4, c^5$ | 88.57 | 88.45 | 86.30 | 91.09 | 87.03 | 90.98 | 87.81 | 91.78 | 91.87 |
| $c^2, c^3, c^4, c^5$ | 86.07 | 86.11 | 85.87 | 86.25 | 86.26 | 86.13 | 85.93 | 86.66 | 86.80 |
| $c^1, c^2, c^3, c^4, c^5$ | 88.81 | 88.54 | 86.63 | 91.33 | 86.59 | 91.21 | 88.10 | 92.21 | 92.33 |

Table 3: Classification accuracy of texture images using different combination methods (2 textures)

The best $ERR$ values of WS, FI, DS0, DS1 and DS2 are determined according to Eq. 21. Since WS has been widely used in the literature, and it outperforms other conventional methods (Av, Md, Mx, and MV), as observed in Table 2, then we will use it as a representative of the conventional methods when performing the comparison with FI, DS0, DS1 and DS2. Figure 3a shows the $ERR$ values when 2 classes are considered. It is clear that the maximum $ERR$ values of these five combination methods are very close, ranging between 14% to 16%. They are obtained by combining the best two classifiers for WS, FI and DS0, while DS1 and DS2 use three classifiers to obtain their maximum $ERR$. As mentioned earlier, The performance of the first four individual classifiers is weaker than that of the $En$. Notice that, for both DS1 and DS2, there is no significant degradation in $ERR$ as the number of combined classifiers increases.

For the case of 3 classes, both DS1 and DS2 outperform other combination methods in terms of the maximum $ERR$. They achieve values of 17.3% and 19.6% respectively, compared to 11.4% or less for other methods as shown in Figure 3b. In addition, $ERR$ of DS1 and DS2 are not affected as the number of combined classifiers increases.

For the case of 5 classes, the maximum $ERR$ values sorted in a descending order are: DS1 50.7%, DS2 40.2%, FI 31.6%, WS 28.1%, and DS0 23.8%, as shown in Figure 3c. In addition, $ERR$ values of DS1 improve as the number of combined classifiers increases, DS2 is the second best, while $ERR$ values of other methods degrade as the number of combined classifiers increases. For the case of 9 classes, the superiority of DS1 becomes clearer, where as shown in Figure 3d, the maximum $ERR$ value of DS1 is 54% compared to 37.5% or less for other methods. It is worth mentioning that even though the maximum $ERR$ values of other methods degrade, they still perform better than the best single classifier. This leads us to conclude that as the number of classes increases, the performance of most classifier combination methods gets better overall.

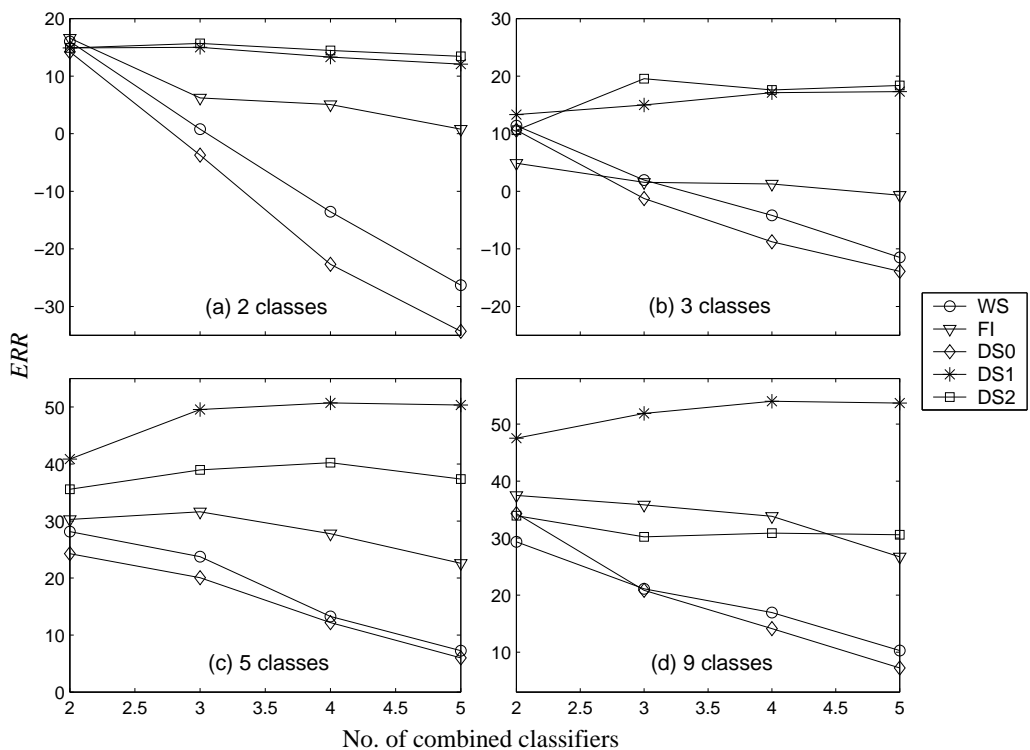

Figure 3: *ERR* of different classifier combination methods obtained by considering different number of classifiers for the cases: (a) 2 classes, (b) 3 classes, (c) 5 classes, and (d) 9 classes

Taking all these facts into consideration, we can sort the methods in a descending order as follows: DS1, DS2, FI, WS, DS0, and the other conventional methods. Thus, in summary, for the problem of texture classification, our proposed technique with its two alternatives (DS1 and DS2) clearly outperforms other standard combination methods with an increase in classification accuracy of about $2-7\%$. For the cases of 2 and 3 classes, there is a little difference in performance between DS1 and DS2. This is because using reference *vectors* of small size, $2 \times 1$ and $3 \times 1$, does not make a big impact upon the estimation of evidence compared to that obtained using a single reference *value*. As the size of the reference vector increases, $5 \times 1$ and $9 \times 1$ for the other two cases, its impact on estimating the evidence becomes clearer, which leads to better results, but at the cost of increasing computational load.

## 5.2 Speech Segment Classification

Six different input feature sets have been used to classify speech segments according to their manner of articulation, these were: 13 mel-frequency cepstral coefficients (MFC), 16 log mel-filter bank (MFB), 12 linear predictive cepstral coefficients (LPC), 12 linear predictive reflection coefficients (LPR), 10 wavelet energy bands (WVT), and 12 autoregressive model parameters (ARP). For this experiment, speech was obtained from the TIMIT database (MIT, SRI, & TI, 1990). Segments of 152 speakers (56456 segments) were used to train the ANNs, and 52 speakers (19228 segments) to test them. Three cases were considered: 3 classes (vowel, consonant, and silence), 6 classes (vowel, nasal, fricative, stop, glide, and silence), and finally 9 classes (vowel, semi-vowel, nasal, fricative, stop, closure, lateral, rhotic, and silence). The classification results for these three cases are summarized in Table 4.

| No. of classes | MFC | MFB | LPC | LPR | WVT | ARP |
|:---:|:---:|:---:|:---:|:---:|:---:|:---:|
| 3 | 88.21 | 90.98 | 81.64 | 80.69 | 90.64 | 70.87 |
| 6 | 83.16 | 85.50 | 74.77 | 74.06 | 84.33 | 62.90 |
| 9 | 78.48 | 83.24 | 71.64 | 70.03 | 81.33 | 56.66 |

Table 4: Speech segment classification accuracy of the six original classifiers for different number of class labels

| No. of classes | WS | Av | Md | Mx | MV | FI | DS0 | DS1 | DS2 |
|:---:|:---:|:---:|:---:|:---:|:---:|:---:|:---:|:---:|:---:|
| 3 | 90.80 | 90.41 | 90.20 | 86.15 | 89.51 | 90.65 | 90.90 | 91.57 | 91.31 |
| 6 | 85.54 | 84.91 | 84.62 | 81.16 | 84.03 | 85.29 | 85.18 | 87.18 | 86.37 |
| 9 | 83.05 | 82.31 | 81.93 | 75.63 | 81.00 | 82.73 | 82.86 | 85.20 | 84.22 |

Table 5: Overall performance of the various combination methods for different number of class labels (speech segment classification)

The two best individual classifiers are MFB and WVT in all three cases, followed by MFC then other methods. Unlike texture classifiers that had one good classifier and four, relatively, weak classifiers, we have here three good classifiers (MFB, MFC and WVT) and three weak classifiers (LPC, LPR and ARP).

The overall performance values of the various combination methods are displayed in Table 5. For the case of the 3 classes, it can be seen that the overall performance of DS1 is better than that of DS2 and they both outperform the other methods. This becomes even clearer as the number of classes increases (with more than 2% increase in accuracy).

The *ERR* values for the case of 3 classes are shown in Figure 4a. The maximum *ERR* value of DS1 is 23.4%, which is achieved by combining all six classifiers, compared to 20.3% for DS2 and 19.6% or less for the other methods. The gap between DS1 and the other methods gets wider when we consider 6 and 9 classes as shown in Figures 4b and 4c. Because there are more good classifiers in this experiment compared to that of the texture experiment, the variations of the *ERR* values when the number of classifiers increases are found to be smaller. In addition, we can see that as the number classes increases DS1 keeps its steady and superior performance in terms of *ERR* with more than 10% increase.

As a summary, DS1 outperforms other methods in terms of overall performance and *ERR* measurements. It is followed by DS2, WS, and the rest of the methods.

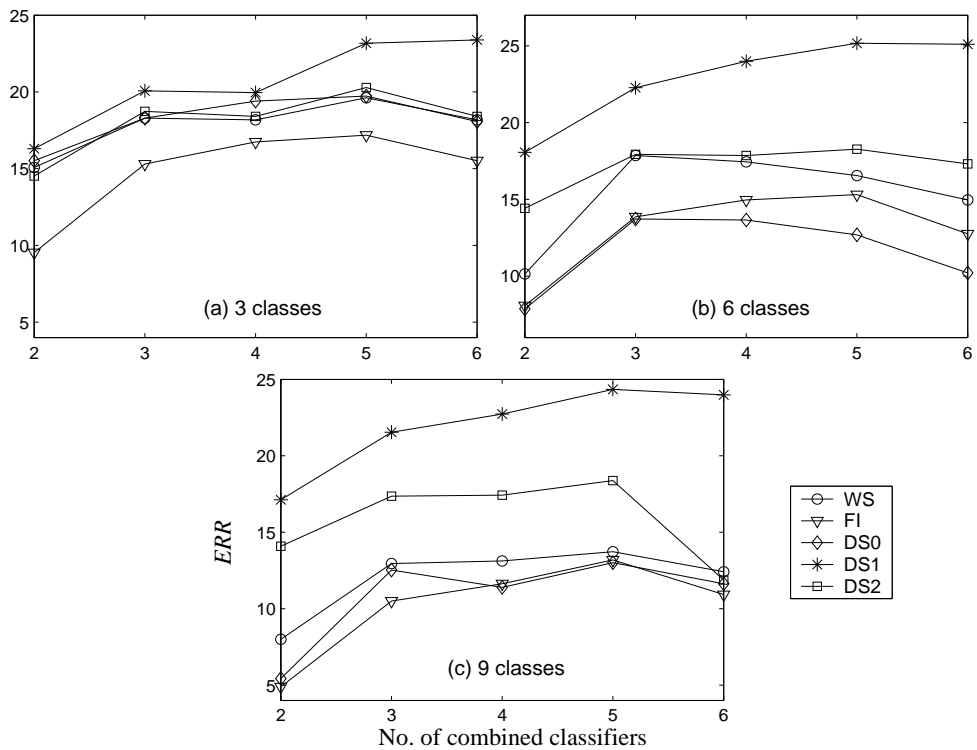

Figure 4: *ERR* of different classifier combination methods obtained by considering different number of classifiers for the cases: (a) 3 classes, (b) 6 classes, and (c) 9 classes

## 5.3 Speaker Identification

Three limited-scope experiments were carried out to perform speaker identification using 2, 3, and 4 speakers. Speech data from the TIMIT database was also used (MIT et al., 1990). The number of training patterns were 3232, 4481 and 5931 respectively, and the number of testing patterns were 1358, 1921 and 2542 respectively. The same features used to classify speech segments according to their manner of articulation were used to identify speakers. Classification results of the six classifiers are shown in Table 6. The performance of the individual classifiers are not quite similar to the speech segment problem, where the three good classifiers are: MFB, MFC and LPC and the three weak classifiers are: LPR, WVT and ARP.

The overall performance of the various combination methods are shown in Table 7. For the case of 2 classes, it is clear that the overall performance of most combination methods is very comparable. The superiority of DS1, and to a lesser degree DS2, becomes clear as the number of classes increases (more patterns were included to estimate evidence).

Note that, because of the high performance of individual classifiers for the case of 2 classes, a small difference in the performance of combination methods will have great impact on *ERR*, which explains the graphs' fluctuations, as shown in Figure 5a. It can be seen that both maximum $ERR$ and overall performance of most combination methods are close. These results do not favor DS1 nor DS2, because they have an additional computational cost. Let's now consider the case of 3 classes, Figure 5b shows that the maximum $ERR$ of DS2 is the highest followed by DS1, and they both outperform the other methods. For the case of 4 classes, the maximum $ERR$ of DS1 is 30%, compared to 27% or less for other methods, as shown in Figure 5c. The figure also shows that $ERR$ values of DS2 and WS are close. However, as the overall performance of DS2 is better than that of WS, DS2 can be considered as the second best method followed by WS, DS0 and finally FI.

| No. of classes | MFC | MFB | LPC | LPR | WVT | ARP |
|---|---|---|---|---|---|---|
| 2 | 94.58 | 96.17 | 92.49 | 89.60 | 87.80 | 84.55 |
| 3 | 85.84 | 87.25 | 82.20 | 81.00 | 74.39 | 73.03 |
| 4 | 85.01 | 85.96 | 80.84 | 77.97 | 70.93 | 64.59 |

Table 6: Speaker identification accuracy of the six original classifiers with different number of speakers

| No. of classes | WS | Av | Md | Mx | MV | FI | DS0 | DS1 | DS2 |
|---|---|---|---|---|---|---|---|---|---|
| 2 | 95.53 | 95.50 | 95.26 | 95.36 | 95.21 | 95.25 | 95.46 | 95.48 | 95.45 |
| 3 | 90.80 | 90.41 | 90.20 | 86.15 | 89.51 | 90.65 | 90.90 | 91.57 | 91.31 |
| 4 | 83.05 | 82.31 | 81.93 | 75.63 | 81.00 | 82.73 | 82.86 | 85.20 | 84.22 |

Table 7: Overall performance of the various combination methods for different number of class labels (speaker identification)

The above results clearly show how the performance of DS1 and DS2 get affected by the number of training patterns, which is crucial in achieving good estimation of the evidence of each classifier. This is very clear for the case of 2 speakers. Their performance, however, get better as the number of speakers and training patterns increase. In other words, DS1 and DS2 require a larger number of patterns to work properly. Failing to provide such number of patterns, other conventional methods, such as WS, can achieve similar performance.

The experiments of textures, speech segments and speaker classification show that our proposed technique clearly outperforms the other methods in terms of overall performance and $ERR$, providing that a sufficient number of patterns to estimate evidence of classifiers exists. Also, among the different combination methods, DS1 and DS2 are the least effected by the inclusion of weak classifiers. The experiments also show that the BBAs could be better estimated using reference *vectors* rather than reference *values*, especially for large number of classes.

It is worth mentioning that each one of the combination methods has its own merit. For example, the MV is very useful combination method when dealing with classifiers that produce results of the abstract level. When working in the measurement level, other combination methods could have better performance.

The Mx method can provide good results when the performance of the combined classifiers are close. In such case, the classifier with higher *confidence* can provide better results

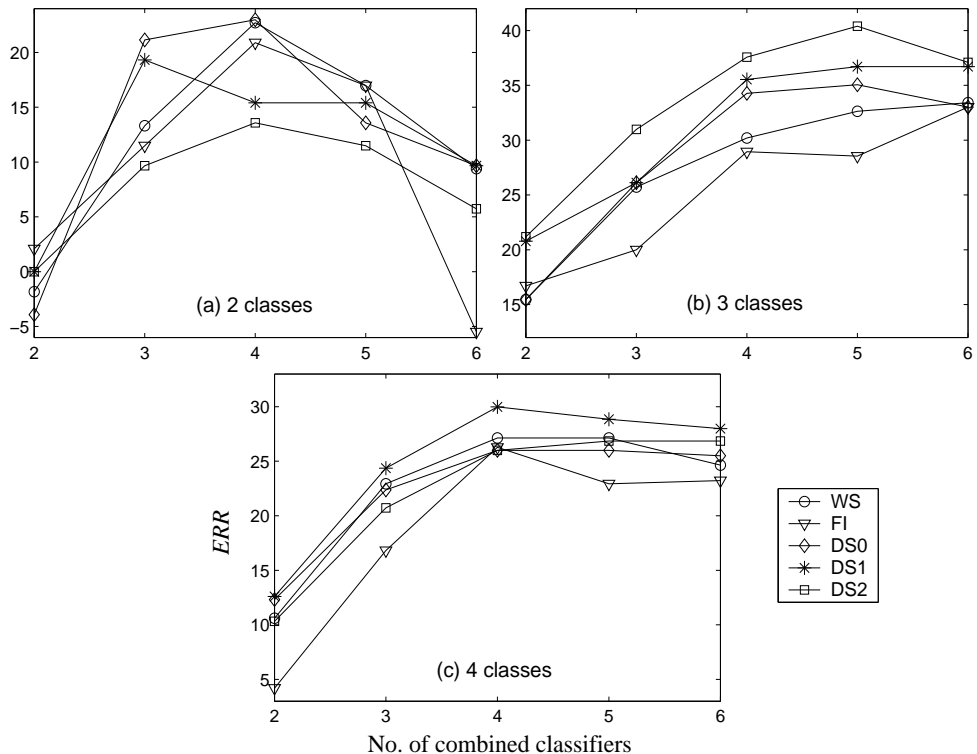

Figure 5: *ERR* of different classifier combination methods obtained by considering different number of classifiers for the cases: (a) 2 classes, (b) 3 classes, and (c) 4 classes

than any individual classifier. This is shown in Tables 11-13 (refer to Appendix B), where good results are achieved when combining the best two or three classifiers of the speech segment experiment compared to the best individual classifier. However, if there is a clear difference in the performance of classifiers, as in the case when considering mixtures of good and bad classifiers, then using Mx to combine the classification results will not be a good choice. In case we don't have any information about the performance of the classifiers, *i.e.*, there is no training dataset, the Av and Md methods could provide an attractive choice. Similar to the findings of (Kittler et al., 1998; Alkoot & Kittler, 1999), the performance of these two methods are found to be close with slight favor of the Av method. If the classification accuracy of the different classifiers are available, then the WS method represents a good choice, where it outperforms Av in almost all the conducted experiments. This is expected, as associating each classifier with a weight that reflects its performance, would make the better classifier contributes more towards the final decision. If the performance of the combined classifiers are very close, then combining their results using both the Av and WS methods would lead to very similar performance, as shown in Tables 11-13 for the cases of combining the best two and three speech segment classifiers.

The FI and DS0 represent two non-linear combination methods. According to (Cho & Kim, 1995a), the performance of FI was slightly better than the WS when tested using an optical character recognition database, which is similar to the results we obtained for the texture experiments. However, for speech segment classification and speaker identification experiments, the performance of FI was not as good as that of WS. On the other hand, the experiments conducted here show that WS slightly outperforms DS0. Note that Rogova (1994) only compared DS0 to the original classifiers. The main problem with both FI and DS0 is the appropriate estimation of their parameters. For example, the desired sum of fuzzy densities affects the combination results of FI, while the choice of the proximity measure and reference vector plays an important role in the performance of DS0.

DS1 and DS2 differ from DS0 by the appropriate measure of the reference vectors, and hence the accurate estimation of the evidence of each classifier. This will exploit the complementary information provided by the different classifiers. In other words, the accurate estimation of evidence of each classifier will lead to minimizing the MSE of the combined results, and hence resolving the conflicts between classifiers.

## 6. Conclusion

We have developed in this work a new powerful classifier combination technique based on the D-S theory of evidence. The technique, based on adjusting the evidence of different classifiers by minimizing the MSE of training data, gave very good results in terms of overall performance and error reduction rate. To test the algorithm, three experiments were carried out: texture classification, speech segments classification, and speaker identification. All of the experiments showed the superiority of the proposed technique when compared to conventional methods, fuzzy integral, and another D-S implementation that uses a different measure of evidence. We have shown that accurate estimation of the evidence from different classifiers based on the whole output vectors (DS1) gives the best performance, especially for higher number of class labels. The only drawback of the algorithm is that training can be computationally expensive (this is used to accurately estimate the evidence of each classifier). However, this can be executed off-line, and as such, has no major effect on the performance of the algorithm. We have also shown that the proposed algorithm can easily achieve an increase in classification accuracy of the order of 2% to 7% compared to other combination methods. We believe that with more work on enhancing the technique, the scheme can form a new framework for pattern classification in the future.

## Acknowledgment

The authors wish to thank Dr. J. Chebil and Dr. M. Mesbah for their valuable comments on the paper. The authors also acknowledge the support of Queensland University of Technology for the work presented in this paper. Dr. Deriche acknowledges the support of King Fahd University, Saudi Arabia, where he is currently on leave.

## Appendix A. Classifier Combination Based on the Fuzzy Integral

Fuzzy integral is a non-linear combination method defined with respect to a fuzzy measure. Detailed explanation of classifier combination based on the $g_\lambda$ fuzzy measure can be found in the work of Cho & Kim (1995a, 1995b).

For a finite set of elements, $Z$, the $g_\lambda$ fuzzy measure (Sugeno, 1977) is defined as the set function $g: 2^Z \to [0, 1]$ that satisfies the following conditions:

1. $g(\emptyset) = 0, g(Z) = 1$,

2. $g(A) \leq g(B)$ if $A \subset B$,

3. if $\{A_i\}_{i=1}^\infty$ is an increasing sequence of measurable sets, then $\lim_{i\to\infty} g(A_i) = g(\lim_{i\to\infty} A_i)$,

4. $g(A \cup B) = g(A) + g(B) + \lambda g(A)g(B)$

for all $A, B \subset Z$ and $A \cap B = \emptyset$, and for some $\lambda > -1$. Let $h : Z \to [0, 1]$ be a fuzzy subset of $Z$. The fuzzy integral over $Z$ of the function $h$ with respect to a fuzzy measure $g$ is defined by

$$\begin{aligned}
h(z) \circ g(.) &= \max_{E \subseteq Z} \left[ \min \left( \min_{z \in E} h(z).g(E) \right) \right] \\
&= \max_{\alpha \in [0,1]} [min(\alpha, g(F_\alpha))], \quad \text{where} \\
F_\alpha &= \{z | h(z) \geq \alpha\}
\end{aligned}$$

Let $Z = \{z_1, \cdots z_n\}$, and suppose that $h(z_1) \geq h(z_2) \geq \cdots \geq h(z_n)$, (if not, $Z$ is rearranged so that this relation holds). Then a fuzzy integral $e$, with respect to a fuzzy measure $g$ over $Z$ can be computed by

$$\begin{aligned}
e &= \max_{i=1}^n [\min(h(z_i), g(A_i))], \quad \text{where} \\
A_i &= \{z_1, \cdots z_i\} \\
g(A_1) &= g(\{z_1\}) = g^1 \\
g(A_i) &= g^i + g(A_{i-1}) + \lambda g^i g(A_{i-1}), \quad \text{for } 1 < i \leq n
\end{aligned}$$

$\lambda$ is given by solving: $\lambda + 1 = \prod_{i=1}^n (1 + \lambda g^i)$, where $\lambda \in (-1, \infty)$ and $\lambda \neq 0$. This can be calculated by solving an $(n-1)^{st}$ degree polynomial and finding the unique root greater than $-1$.

For the problem of combining classifiers, $Z$ represents the set of classifiers, $A$ the object under consideration for classification, and $h_k(z_i)$ is the partial evaluation of the object $A$ for class $\omega_k$. Corresponding to each classifier $z_i$, the degree of importance, $g^i$, that reflects how good is $z_i$ in the classification of class $\omega_k$ must be given. These densities can be induced from a training dataset.

## Appendix B. Tables of Classification Accuracy for Different Combination Methods

| Classifiers | WS | Av | Md | Mx | MV | FI | DS0 | DS1 | DS2 |
|---|---|---|---|---|---|---|---|---|---|
| $c^1, c^2$ | **90.89** | **90.79** | **90.79** | **90.83** | **90.17** | **90.22** | **90.81** | 91.09 | 90.81 |
| $c^1, c^5$ | 89.69 | 89.57 | 89.57 | 89.68 | 89.10 | 88.99 | 89.92 | 90.45 | 90.29 |
| $c^4, c^5$ | 85.05 | 85.05 | 85.05 | 84.67 | 84.81 | 85.13 | 85.51 | 85.49 | 85.44 |
| $c^1, c^2, c^3$ | 89.92 | 89.83 | 88.04 | 90.48 | 87.77 | 89.88 | 89.59 | 91.26 | **91.73** |
| $c^1, c^2, c^5$ | 89.55 | 89.39 | 87.54 | 90.23 | 87.09 | 89.56 | 89.18 | 91.12 | 91.10 |
| $c^1, c^4, c^5$ | 89.05 | 88.80 | 86.95 | 89.62 | 86.62 | 89.26 | 88.80 | 90.77 | 90.73 |
| $c^3, c^4, c^5$ | 85.76 | 85.73 | 85.49 | 84.81 | 85.49 | 85.67 | 85.91 | 87.27 | 86.88 |
| $c^1, c^2, c^3, c^4$ | 89.29 | 89.07 | 87.98 | 90.17 | 87.91 | 89.85 | 88.82 | 91.44 | 91.51 |
| $c^1, c^2, c^3, c^5$ | 89.07 | 88.83 | 87.24 | 90.01 | 87.50 | 89.58 | 88.81 | 91.48 | 91.53 |
| $c^1, c^2, c^4, c^5$ | 88.87 | 88.78 | 87.34 | 90.02 | 87.61 | 89.64 | 88.66 | 91.16 | 91.26 |
| $c^1, c^3, c^4, c^5$ | 88.60 | 88.38 | 87.05 | 89.39 | 87.52 | 89.44 | 88.39 | 91.31 | 91.00 |
| $c^2, c^3, c^4, c^5$ | 86.45 | 86.44 | 86.18 | 85.52 | 86.31 | 86.36 | 86.46 | 88.37 | 87.20 |
| $c^1, c^2, c^3, c^4, c^5$ | 88.54 | 88.47 | 87.06 | 89.73 | 87.03 | 89.65 | 88.29 | **91.50** | 91.61 |

Table 8: Classification accuracy of texture images using different combination methods (3 textures)

| Classifiers | WS | Av | Md | Mx | MV | FI | DS0 | DS1 | DS2 |
|---|---|---|---|---|---|---|---|---|---|
| $c^1, c^2$ | **91.98** | **91.87** | **91.87** | 90.28 | **89.89** | **92.21** | 91.45 | 93.40 | 92.81 |
| $c^1, c^5$ | 91.95 | 91.74 | 91.74 | **90.72** | 89.58 | 91.30 | **91.55** | 93.27 | 92.72 |
| $c^4, c^5$ | 85.14 | 85.19 | 85.19 | 84.77 | 84.50 | 84.94 | 85.82 | 86.28 | 85.36 |
| $c^1, c^2, c^3$ | 91.49 | 91.29 | 88.78 | 90.39 | 88.60 | 92.35 | 91.08 | 94.37 | 93.19 |
| $c^1, c^2, c^5$ | 90.77 | 90.50 | 87.92 | 90.34 | 87.74 | 91.54 | 90.32 | 93.66 | 92.96 |
| $c^1, c^4, c^5$ | 90.61 | 90.35 | 87.67 | 90.29 | 87.43 | 91.41 | 90.25 | 93.69 | 92.90 |
| $c^3, c^4, c^5$ | 86.29 | 86.22 | 85.40 | 85.99 | 85.64 | 85.91 | 86.98 | 89.16 | 87.29 |
| $c^1, c^2, c^3, c^4$ | 90.26 | 90.03 | 88.37 | 90.12 | 88.90 | 91.86 | 89.93 | 94.48 | 93.19 |
| $c^1, c^2, c^3, c^5$ | 90.14 | 89.91 | 88.17 | 90.25 | 89.00 | 91.83 | 90.09 | 94.41 | 93.26 |
| $c^1, c^2, c^4, c^5$ | 89.73 | 89.43 | 87.62 | 90.19 | 88.45 | 91.11 | 89.41 | 93.82 | 92.56 |
| $c^1, c^3, c^4, c^5$ | 90.32 | 90.06 | 87.91 | 90.50 | 88.51 | 91.83 | 90.20 | **94.50** | **93.33** |
| $c^2, c^3, c^5, c^6$ | 86.41 | 86.38 | 85.99 | 85.98 | 85.79 | 86.15 | 87.21 | 89.47 | 86.97 |
| $c^1, c^2, c^3, c^4, c^5$ | 89.65 | 89.39 | 87.27 | 90.11 | 87.81 | 91.22 | 89.51 | 94.46 | 93.01 |

Table 9: Classification accuracy of texture images using different combination methods (5 textures)

| Classifiers | WS | Av | Md | Mx | MV | FI | DS0 | DS1 | DS2 |
|---|---|---|---|---|---|---|---|---|---|
| $c^1, c^2$ | **88.45** | **88.13** | **88.13** | **85.44** | **85.91** | **89.78** | **89.26** | 91.42 | **89.20** |
| $c^1, c^5$ | 86.39 | 85.97 | 85.97 | 82.27 | 83.00 | 88.18 | 87.10 | 90.07 | 88.08 |
| $c^4, c^5$ | 79.55 | 79.40 | 79.40 | 78.74 | 78.53 | 79.37 | 79.28 | 81.22 | 79.91 |
| $c^1, c^2, c^3$ | 87.10 | 86.53 | 84.20 | 84.65 | 84.60 | 89.51 | 87.06 | 92.00 | 88.59 |
| $c^1, c^2, c^5$ | 86.48 | 85.85 | 83.66 | 84.08 | 84.17 | 88.72 | 86.92 | 92.13 | 88.53 |
| $c^1, c^4, c^5$ | 85.92 | 85.32 | 83.22 | 82.97 | 83.67 | 88.25 | 85.98 | 92.01 | 88.30 |
| $c^3, c^4, c^5$ | 80.27 | 80.21 | 79.63 | 79.54 | 79.38 | 79.97 | 79.80 | 82.48 | 80.89 |
| $c^1, c^2, c^3, c^4$ | 86.42 | 85.95 | 84.56 | 84.33 | 85.16 | 89.18 | 85.96 | 92.45 | 88.70 |
| $c^1, c^2, c^3, c^5$ | 85.70 | 85.11 | 83.37 | 83.91 | 84.26 | 88.52 | 85.41 | 92.02 | 88.30 |
| $c^1, c^2, c^4, c^5$ | 85.79 | 85.39 | 84.29 | 83.83 | 84.91 | 88.52 | 85.94 | **92.48** | 88.66 |
| $c^1, c^3, c^4, c^5$ | 85.42 | 84.92 | 83.06 | 83.47 | 84.08 | 88.34 | 84.81 | 92.35 | 88.35 |
| $c^2, c^3, c^4, c^5$ | 81.60 | 81.44 | 81.15 | 80.75 | 81.03 | 81.54 | 81.02 | 84.70 | 82.17 |
| $c^1, c^2, c^3, c^4, c^5$ | 85.33 | 84.88 | 83.19 | 83.74 | 83.35 | 88.02 | 84.83 | 92.43 | 88.65 |

Table 10: Classification accuracy of texture images using different combination methods (9 textures)

| Classifiers | WS | Av | Md | Mx | MV | FI | DS0 | DS1 | DS2 |
|---|---|---|---|---|---|---|---|---|---|
| $c^1, c^2$ | 92.34 | 92.34 | 92.34 | 92.34 | 92.22 | 91.84 | 92.38 | 92.45 | 92.29 |
| $c^1, c^6$ | 89.99 | 88.11 | 88.11 | 83.95 | 82.64 | 91.06 | 90.92 | 91.48 | 91.42 |
| $c^5, c^6$ | 81.09 | 80.65 | 80.65 | 76.19 | 76.10 | 82.03 | 81.54 | 82.29 | 82.01 |
| $c^1, c^2, c^3$ | 92.63 | 92.61 | 92.37 | **92.34** | **92.27** | 92.36 | 92.63 | 92.79 | 92.67 |
| $c^1, c^2, c^6$ | 92.17 | 91.79 | 91.83 | 85.97 | 91.60 | 92.03 | 92.27 | 92.53 | 92.23 |
| $c^1, c^5, c^6$ | 89.85 | 88.79 | 88.20 | 84.17 | 87.50 | 89.66 | 90.34 | 91.92 | 91.79 |
| $c^4, c^5, c^6$ | 84.98 | 84.96 | 84.76 | 79.36 | 84.08 | 84.89 | 84.62 | 85.65 | 85.36 |
| $c^1, c^2, c^3, c^4$ | 92.59 | 92.57 | 92.42 | 91.64 | 92.04 | 92.38 | 92.61 | 92.77 | 92.64 |
| $c^1, c^2, c^3, c^6$ | 92.62 | 92.47 | **92.50** | 86.94 | 92.02 | 92.49 | 92.73 | 92.78 | 92.54 |
| $c^1, c^2, c^5, c^6$ | 92.25 | 91.93 | 91.82 | 86.01 | 91.96 | 92.14 | 92.37 | 92.77 | 92.49 |
| $c^1, c^4, c^5, c^6$ | 90.15 | 89.51 | 89.34 | 84.64 | 89.48 | 89.48 | 90.12 | 91.92 | 91.84 |
| $c^3, c^4, c^5, c^6$ | 88.79 | 88.42 | 88.30 | 83.25 | 88.46 | 88.46 | 88.76 | 90.51 | 89.94 |
| $c^1, c^2, c^3, c^4, c^5$ | **92.75** | **92.64** | 92.23 | 91.42 | 92.13 | 92.49 | 92.74 | 93.07 | **92.81** |
| $c^1, c^2, c^3, c^4, c^6$ | 92.57 | 92.48 | 92.30 | 86.96 | 91.92 | 92.25 | 92.56 | 92.77 | 92.56 |
| $c^1, c^2, c^3, c^5, c^6$ | 92.73 | 92.50 | 92.20 | 86.93 | 91.91 | **92.53** | **92.76** | 93.03 | 92.72 |
| $c^1, c^2, c^4, c^5, c^6$ | 92.14 | 91.90 | 91.07 | 86.19 | 90.97 | 91.71 | 92.23 | 92.78 | 92.46 |
| $c^1, c^3, c^4, c^5, c^6$ | 91.41 | 91.04 | 90.63 | 85.79 | 90.41 | 90.98 | 91.46 | 92.48 | 92.14 |
| $c^2, c^3, c^4, c^5, c^6$ | 91.51 | 90.98 | 90.52 | 85.73 | 90.60 | 91.15 | 91.50 | 92.67 | 92.40 |
| $c^1, c^2, c^3, c^4, c^5, c^6$ | 92.62 | 92.36 | 92.24 | 86.95 | 91.97 | 92.38 | 92.61 | **93.09** | 92.64 |

Table 11: Classification accuracy of speech segments using different combination methods (3 classes)

| Classifiers | WS | Av | Md | Mx | MV | FI | DS0 | DS1 | DS2 |
|---|---|---|---|---|---|---|---|---|---|
| $c^1, c^2$ | 86.97 | 86.95 | 86.95 | 86.46 | 86.30 | 86.67 | 86.64 | 88.12 | 87.59 |
| $c^1, c^6$ | 84.30 | 81.98 | 81.98 | 77.97 | 76.66 | 85.01 | 84.52 | 86.71 | 86.39 |
| $c^5, c^6$ | 74.60 | 73.91 | 73.91 | 70.38 | 70.20 | 75.00 | 74.20 | 76.47 | 75.68 |
| $c^1, c^2, c^3$ | **88.09** | **88.08** | 87.77 | **87.11** | **87.76** | 87.51 | **87.49** | 88.73 | 88.10 |
| $c^1, c^2, c^6$ | 86.48 | 85.76 | 86.01 | 80.38 | 85.31 | 86.49 | 86.45 | 88.19 | 87.50 |
| $c^1, c^5, c^6$ | 84.33 | 82.80 | 81.88 | 78.64 | 81.24 | 83.88 | 84.56 | 87.16 | 86.45 |
| $c^4, c^5, c^6$ | 79.27 | 78.73 | 77.84 | 74.07 | 77.26 | 78.46 | 78.51 | 80.17 | 79.39 |
| $c^1, c^2, c^3, c^4$ | 88.03 | 87.91 | **87.81** | 86.58 | 87.29 | 87.67 | 87.48 | 88.98 | 88.09 |
| $c^1, c^2, c^3, c^6$ | 87.64 | 87.17 | 87.26 | 82.59 | 86.95 | 87.37 | 87.25 | 88.61 | 88.08 |
| $c^1, c^2, c^5, c^6$ | 86.51 | 85.84 | 85.90 | 80.54 | 86.10 | 86.69 | 86.59 | 88.47 | 87.57 |
| $c^1, c^4, c^5, c^6$ | 84.61 | 83.64 | 83.28 | 79.47 | 83.85 | 83.97 | 84.29 | 87.38 | 86.68 |
| $c^3, c^4, c^5, c^6$ | 83.84 | 83.12 | 82.84 | 79.28 | 83.05 | 83.22 | 83.56 | 86.00 | 84.95 |
| $c^1, c^2, c^3, c^4, c^5$ | 87.90 | 87.87 | 87.44 | 86.01 | 87.35 | **87.72** | 87.34 | **89.15** | **88.15** |
| $c^1, c^2, c^3, c^4, c^6$ | 87.57 | 87.24 | 86.98 | 82.76 | 86.85 | 87.36 | 87.08 | 88.95 | 88.11 |
| $c^1, c^2, c^3, c^5, c^6$ | 87.69 | 87.17 | 86.91 | 82.64 | 86.96 | 87.52 | 87.32 | 88.93 | 88.12 |
| $c^1, c^2, c^4, c^5, c^6$ | 86.71 | 86.00 | 85.54 | 81.21 | 85.63 | 86.46 | 86.39 | 88.64 | 87.56 |
| $c^1, c^3, c^4, c^5, c^6$ | 86.36 | 85.85 | 85.26 | 81.50 | 85.57 | 86.10 | 85.95 | 88.26 | 87.38 |
| $c^2, c^3, c^4, c^5, c^6$ | 86.70 | 85.95 | 85.18 | 81.63 | 85.23 | 86.06 | 85.81 | 88.37 | 87.28 |
| $c^1, c^2, c^3, c^4, c^5, c^6$ | 87.67 | 87.33 | 87.02 | 82.76 | 86.95 | 87.35 | 86.98 | 89.14 | 88.01 |

Table 12: Classification accuracy of speech segments using different combination methods (6 classes)

| Classifiers | WS | Av | Md | Mx | MV | FI | DS0 | DS1 | DS2 |
|---|---|---|---|---|---|---|---|---|---|
| $c^1, c^2$ | 84.58 | 84.61 | 84.61 | 83.77 | 83.66 | 84.06 | 84.15 | 86.11 | 85.60 |
| $c^1, c^6$ | 81.85 | 78.59 | 78.59 | 71.48 | 70.50 | 82.53 | 82.51 | 84.41 | 83.72 |
| $c^5, c^6$ | 70.48 | 69.04 | 69.04 | 62.87 | 62.65 | 71.22 | 69.97 | 73.24 | 71.94 |
| $c^1, c^2, c^3$ | 85.41 | 85.42 | 85.00 | **83.81** | **85.02** | 85.00 | 85.34 | 86.85 | 86.15 |
| $c^1, c^2, c^6$ | 84.20 | 83.44 | 83.53 | 74.18 | 82.70 | 84.11 | 84.01 | 86.05 | 85.44 |
| $c^1, c^5, c^6$ | 82.18 | 80.26 | 79.19 | 72.32 | 77.86 | 81.20 | 82.53 | 85.10 | 84.56 |
| $c^4, c^5, c^6$ | 75.94 | 75.23 | 74.43 | 67.23 | 73.46 | 75.30 | 75.54 | 78.21 | 76.92 |
| $c^1, c^2, c^3, c^4$ | 85.44 | 85.30 | **85.20** | 83.41 | 84.96 | 85.11 | 85.15 | 87.05 | 86.13 |
| $c^1, c^2, c^3, c^6$ | 85.34 | 84.97 | 84.99 | 76.49 | 84.82 | 85.19 | 85.14 | 86.79 | 86.16 |
| $c^1, c^2, c^5, c^6$ | 84.48 | 83.77 | 83.77 | 74.61 | 83.98 | 84.58 | 84.53 | 86.67 | 85.86 |
| $c^1, c^4, c^5, c^6$ | 82.41 | 81.44 | 81.03 | 73.61 | 81.49 | 81.61 | 82.44 | 85.65 | 84.76 |
| $c^3, c^4, c^5, c^6$ | 81.02 | 80.51 | 80.15 | 72.74 | 79.83 | 80.06 | 80.41 | 83.86 | 82.25 |
| $c^1, c^2, c^3, c^4, c^5$ | 85.52 | **85.43** | 84.94 | 82.93 | 84.77 | **85.45** | **85.42** | **87.32** | **86.32** |
| $c^1, c^2, c^3, c^4, c^6$ | 85.53 | 85.01 | 84.41 | 77.04 | 84.53 | 84.96 | 85.00 | 87.09 | 86.09 |
| $c^1, c^2, c^3, c^5, c^6$ | **85.54** | 85.16 | 84.76 | 76.62 | 84.66 | 85.38 | 85.25 | 87.11 | 86.22 |
| $c^1, c^2, c^4, c^5, c^6$ | 84.60 | 84.03 | 83.34 | 75.51 | 83.69 | 84.17 | 84.42 | 86.82 | 85.78 |
| $c^1, c^3, c^4, c^5, c^6$ | 83.97 | 83.32 | 82.74 | 75.80 | 83.16 | 83.67 | 83.91 | 86.59 | 85.45 |
| $c^2, c^3, c^4, c^5, c^6$ | 84.14 | 83.48 | 82.32 | 75.40 | 82.66 | 83.29 | 83.44 | 86.55 | 85.56 |
| $c^1, c^2, c^3, c^4, c^5, c^6$ | 85.32 | 84.93 | 84.70 | 77.08 | 84.51 | 85.07 | 85.19 | 87.26 | 85.23 |

Table 13: Classification accuracy of speech segments using different combination methods (9 classes)

| Classifiers | WS | Av | Md | Mx | MV | FI | DS0 | DS1 | DS2 |
|---|---|---|---|---|---|---|---|---|---|
| $c^1,\ c^2$ | 96.10 | 96.10 | 96.10 | 96.10 | 96.10 | 95.16 | 96.02 | 95.51 | 95.80 |
| $c^1,\ c^6$ | 95.38 | 94.95 | 94.95 | 94.95 | 94.95 | 96.25 | 95.36 | 96.17 | 96.17 |
| $c^5,\ c^6$ | 90.25 | 90.69 | 90.69 | 90.83 | 90.83 | 87.51 | 89.32 | 88.59 | 88.66 |
| $c^1,\ c^2,\ c^3$ | 96.68 | 96.68 | 96.90 | **96.97** | **96.90** | 96.61 | 96.98 | **96.91** | 96.54 |
| $c^1,\ c^2,\ c^6$ | 95.81 | 95.74 | 95.23 | 95.38 | 95.23 | 96.17 | 95.73 | 95.51 | 95.88 |
| $c^1,\ c^5,\ c^6$ | 94.66 | 94.73 | 94.37 | 95.09 | 94.51 | 94.95 | 94.62 | 96.24 | 96.17 |
| $c^4,\ c^5,\ c^6$ | 92.20 | 92.13 | 91.77 | 92.64 | 91.70 | 91.48 | 91.75 | 91.16 | 91.16 |
| $c^1,\ c^2,\ c^3,\ c^4$ | 96.90 | **96.90** | **96.97** | 96.68 | 96.61 | 96.61 | **97.05** | 96.69 | **96.69** |
| $c^1,\ c^2,\ c^3,\ c^6$ | **97.04** | 96.82 | 96.97 | 96.10 | 96.53 | **96.97** | 96.91 | 96.76 | 96.69 |
| $c^1,\ c^2,\ c^5,\ c^6$ | 95.88 | 95.88 | 95.74 | 95.16 | 95.96 | 96.17 | 95.95 | 95.80 | 95.80 |
| $c^1,\ c^4,\ c^5,\ c^6$ | 95.88 | 95.81 | 95.52 | 95.52 | 95.45 | 94.95 | 95.95 | 95.88 | 96.24 |
| $c^3,\ c^4,\ c^5,\ c^6$ | 95.02 | 94.95 | 94.30 | 94.01 | 94.30 | 94.30 | 95.14 | 94.55 | 94.40 |
| $c^1,\ c^2,\ c^3,\ c^4,\ c^5$ | 96.82 | 96.75 | 96.25 | 96.53 | 96.25 | 96.82 | 96.69 | 96.76 | 96.61 |
| $c^1,\ c^2,\ c^3,\ c^4,\ c^6$ | 96.10 | 96.10 | 95.45 | 96.32 | 95.45 | 96.25 | 96.02 | 96.24 | 96.10 |
| $c^1,\ c^2,\ c^3,\ c^5,\ c^6$ | 96.53 | 96.46 | 96.17 | 95.88 | 96.10 | 96.75 | 96.47 | 96.76 | 96.54 |
| $c^1,\ c^2,\ c^4,\ c^5,\ c^6$ | 95.74 | 95.67 | 95.38 | 95.60 | 95.31 | 95.81 | 95.88 | 96.02 | 95.95 |
| $c^1,\ c^3,\ c^4,\ c^5,\ c^6$ | 96.39 | 96.32 | 95.60 | 96.39 | 95.52 | 95.96 | 96.24 | 96.39 | 96.24 |
| $c^2,\ c^3,\ c^4,\ c^5,\ c^6$ | 95.16 | 95.23 | 95.02 | 95.45 | 94.95 | 95.16 | 95.14 | 95.58 | 95.58 |
| $c^1,\ c^2,\ c^3,\ c^4,\ c^5,\ c^6$ | 96.53 | 96.61 | 96.53 | 96.17 | 96.25 | 95.96 | 96.54 | 96.54 | 96.39 |

Table 14: Speaker identification accuracy using different combination methods (2 speakers)

| Classifiers | WS | Av | Md | Mx | MV | FI | DS0 | DS1 | DS2 |
|---|---|---|---|---|---|---|---|---|---|
| $c^1, c^2$ | 89.22 | 89.17 | 89.17 | 88.70 | 88.91 | 87.82 | 89.22 | 89.33 | 89.17 |
| $c^1, c^6$ | 88.91 | 88.55 | 88.55 | 87.61 | 85.58 | 89.38 | 88.86 | 89.90 | 89.95 |
| $c^5, c^6$ | 80.37 | 80.90 | 80.90 | 79.44 | 77.93 | 75.90 | 79.44 | 80.37 | 79.85 |
| $c^1, c^2, c^3$ | 90.53 | 90.47 | 90.32 | 89.28 | 90.06 | 89.80 | 90.47 | 90.58 | 91.20 |
| $c^1, c^2, c^6$ | 90.37 | 90.58 | 89.90 | 88.65 | 90.06 | 89.43 | 90.58 | 90.32 | 90.21 |
| $c^1, c^5, c^6$ | 88.70 | 88.60 | 87.35 | 87.25 | 86.93 | 88.29 | 88.96 | 90.11 | 90.58 |
| $c^4, c^5, c^6$ | 84.54 | 84.17 | 83.60 | 83.24 | 83.34 | 83.08 | 83.86 | 84.64 | 85.06 |
| $c^1, c^2, c^3, c^4$ | 91.10 | 91.10 | 90.58 | **89.59** | 90.32 | 90.94 | 91.31 | 91.78 | 91.51 |
| $c^1, c^2, c^3, c^6$ | 91.25 | 91.51 | 91.36 | 88.70 | 90.06 | 90.89 | 91.62 | 91.62 | 92.04 |
| $c^1, c^2, c^5, c^6$ | 90.63 | 90.47 | 90.47 | 88.81 | 88.39 | 90.16 | 90.58 | 90.94 | 90.21 |
| $c^1, c^4, c^5, c^6$ | 90.37 | 90.32 | 89.48 | 87.87 | 88.55 | 89.54 | 90.47 | 91.57 | 91.46 |
| $c^3, c^4, c^5, c^6$ | 86.88 | 86.78 | 85.94 | 84.64 | 84.54 | 85.79 | 86.15 | 87.40 | 87.35 |
| $c^1, c^2, c^3, c^4, c^5$ | 91.20 | 91.20 | 90.58 | 89.59 | 89.59 | 90.84 | **91.72** | **91.93** | 91.78 |
| $c^1, c^2, c^3, c^4, c^6$ | 91.36 | 91.31 | 91.25 | 88.96 | 90.32 | 90.89 | 91.36 | 91.93 | **92.40** |
| $c^1, c^2, c^3, c^5, c^6$ | 90.99 | 91.15 | 90.58 | 88.81 | 90.06 | 90.73 | 91.20 | 91.62 | 91.88 |
| $c^1, c^2, c^4, c^5, c^6$ | 91.41 | **91.67** | 90.99 | 88.86 | 90.47 | 90.89 | 91.46 | 91.83 | 92.09 |
| $c^1, c^3, c^4, c^5, c^6$ | 91.05 | 90.99 | 89.69 | 88.34 | 89.59 | 89.85 | 91.04 | 91.83 | 91.41 |
| $c^2, c^3, c^4, c^5, c^6$ | 90.37 | 90.27 | 89.22 | 88.24 | 88.81 | 88.96 | 89.38 | 90.42 | 90.32 |
| $c^1, c^2, c^3, c^4, c^5, c^6$ | **91.51** | 91.36 | **91.78** | 88.91 | **90.89** | **91.46** | 91.46 | 91.93 | 91.98 |

Table 15: Speaker identification accuracy using different combination methods (3 speakers)

| Classifiers | WS | Av | Md | Mx | MV | FI | DS0 | DS1 | DS2 |
|---|---|---|---|---|---|---|---|---|---|
| $c^1$, $c^2$ | 87.45 | 87.53 | 87.53 | 87.29 | 86.98 | 86.55 | 87.69 | 87.73 | 87.41 |
| $c^1$, $c^6$ | 84.78 | 83.79 | 83.79 | 83.40 | 81.55 | 85.48 | 83.87 | 85.37 | 84.78 |
| $c^5$, $c^6$ | 74.00 | 74.47 | 74.47 | 72.03 | 71.28 | 69.83 | 71.99 | 73.29 | 73.13 |
| $c^1$, $c^2$, $c^3$ | 89.18 | 89.18 | 88.24 | **88.16** | 87.92 | 88.32 | 89.10 | 89.38 | 88.87 |
| $c^1$, $c^2$, $c^6$ | 87.96 | 88.08 | 87.53 | 86.35 | 87.33 | 87.25 | 87.65 | 88.08 | 87.41 |
| $c^1$, $c^5$, $c^6$ | 85.60 | 85.41 | 83.32 | 83.52 | 83.12 | 84.66 | 84.62 | 86.15 | 85.13 |
| $c^4$, $c^5$, $c^6$ | 80.84 | 80.68 | 79.58 | 77.18 | 78.76 | 80.29 | 78.80 | 81.51 | 80.68 |
| $c^1$, $c^2$, $c^3$, $c^4$ | **89.77** | **89.85** | **89.65** | 88.04 | 88.59 | **89.65** | **89.61** | **90.17** | 89.61 |
| $c^1$, $c^2$, $c^3$, $c^6$ | 89.26 | 89.22 | 88.71 | 87.29 | 88.20 | 88.75 | 88.91 | 89.50 | 88.71 |
| $c^1$, $c^2$, $c^5$, $c^6$ | 87.84 | 87.33 | 86.90 | 86.35 | 87.10 | 87.37 | 87.65 | 88.32 | 87.69 |
| $c^1$, $c^4$, $c^5$, $c^6$ | 87.06 | 86.74 | 86.23 | 83.99 | 85.44 | 86.82 | 86.66 | 88.16 | 87.14 |
| $c^3$, $c^4$, $c^5$, $c^6$ | 84.74 | 84.38 | 84.30 | 80.72 | 83.01 | 84.07 | 83.83 | 85.68 | 84.66 |
| $c^1$, $c^2$, $c^3$, $c^4$, $c^5$ | 89.61 | 89.61 | 89.02 | 88.00 | 88.47 | 89.18 | 89.61 | 90.01 | **89.73** |
| $c^1$, $c^2$, $c^3$, $c^4$, $c^6$ | 89.77 | 89.73 | 88.83 | 87.45 | **88.87** | 89.06 | 89.61 | 89.93 | 89.50 |
| $c^1$, $c^2$, $c^3$, $c^5$, $c^6$ | 89.26 | 89.14 | 88.36 | 87.29 | 88.04 | 88.55 | 89.18 | 89.73 | 88.59 |
| $c^1$, $c^2$, $c^4$, $c^5$, $c^6$ | 89.10 | 89.06 | 88.20 | 86.74 | 88.36 | 88.36 | 88.95 | 89.65 | 88.95 |
| $c^1$, $c^3$, $c^4$, $c^5$, $c^6$ | 88.00 | 87.88 | 86.98 | 85.21 | 87.45 | 87.92 | 87.88 | 89.38 | 89.02 |
| $c^2$, $c^3$, $c^4$, $c^5$, $c^6$ | 88.99 | 88.87 | 87.21 | 85.80 | 87.06 | 88.00 | 87.92 | 89.26 | 88.75 |
| $c^1$, $c^2$, $c^3$, $c^4$, $c^5$, $c^6$ | 89.42 | 89.26 | 88.91 | 87.37 | 88.24 | 89.22 | 89.54 | 89.89 | 89.73 |

Table 16: Speaker identification accuracy using different combination methods (4 speakers)